\documentclass[lettersize,journal]{IEEEtran}
\usepackage{amsmath,amsfonts}
\usepackage{algorithmic}
\usepackage{algorithm}
\usepackage{array}
\usepackage[caption=false,font=normalsize,labelfont=sf,textfont=sf]{subfig}
\usepackage{textcomp}
\usepackage{stfloats}
\usepackage{url}
\usepackage{verbatim}
\usepackage{graphicx}
\usepackage{cite}
\usepackage{booktabs}

\begin{document}

\title{FLoRA: Fusion-Latent for Optical Reconstruction and Flood Area Segmentation via Cross-Modal Multi-Task Distillation Network}

\author{Jagrati Talreja, Tewodros Syum Gebre, Leila Hashemi-Beni

\thanks{This research is funded by the National Science Foundation (NSF) grant 2401942 and by National Aeronautics and Space Administration (NASA). }
\thanks{Jagrati Talreja is with Geomatics Program, College of Science and Technology, North Carolina A\&T Technical State University, Greensboro, North Carolina 27411, USA.}
\thanks{Tewodros Syum Gebre is with Geomatics Program, College of Science and Technology, North Carolina A\&T Technical State University, Greensboro, North Carolina 27411, USA.}
\thanks{Leila Hashemi Beni (corresponding author) is with Geomatics Program, College of Science and Technology, North Carolina A\&T Technical State University, Greensboro, North Carolina 27411, USA. She is also with Institute for Water, Environment and Health, United Nations University, Richmond Hill, Ontario, Canada. \\
E-mail:lhashemibeni@ncat.edu}


}





\maketitle

\begin{abstract}
Accurate flood-water mapping is critical for disaster management, yet current methods struggle to fully exploit the potential of spaceborne imagery. Optical data offers high interpretability but is limited by environmental conditions, whereas SAR provides reliable all-weather coverage with reduced visual interpretability. FLoRA (Fusion-Latent for Optical Reconstruction \& Area Segmentation) is a cross-modal multi-task framework that jointly reconstructs high-fidelity optical imagery and segments flood-water regions from Sentinel-1 SAR by fusing the complementary strengths of optical and SAR data. During training, a lightweight optical teacher (driven by RGB and NDVI priors) provides pyramidal features that guide SAR representations into a fusion-latent space via multi-scale windowed cross-attention and FiLM conditioning, with gated residuals preventing over-correction. This design enables multi-task learning across two complementary objectives: (i) SAR-to-optical translation for fine-grained RGB reconstruction and (ii) flood-water region segmentation for hydrologic interpretation. The dual decoders are optimized using Charbonnier + SSIM for structural fidelity, edge + FFT-magnitude losses for spectral realism, and Dice + BCE + hydrology-aware edge alignment for precise flood-water delineation. A feature-distillation constraint further aligns fused SAR features with the optical teacher’s manifold. Evaluations on SEN1FLOODS11, DEEPFLOOD, and SEN12MS demonstrate that FLoRA surpasses fusion baselines in PSNR, SSIM, and LPIPS, demonstrating that multi-modal fusion within a teacher-guided latent space yields semantically faithful and physically consistent flood-water intelligence from spaceborne observations. \textit{https://github.com/JagratiTalreja01/FLoRA}
\end{abstract}


\section{Introduction}
\label{sec:intro}

Floods remain among the most frequent and devastating natural disasters, responsible for extensive economic losses and significant impacts on human life and infrastructure \cite{jongman2014increasing}. Rapid and reliable flood-water mapping from satellite imagery is critical for early response, damage assessment, and long-term mitigation planning \cite{schumann2018need}. However, the remote sensing modalities for this task are optical imagery \cite{chini2019sentinel, drusch2012sentinel} and synthetic aperture radar (SAR) \cite{torres2012gmes}. Each suffers from inherent limitations \cite{pulvirenti2011flood}. Optical sensors, such as those aboard Sentinel-2, provide visually interpretable reflectance information, but their usability degrades sharply under heavy cloud cover or nighttime conditions \cite{ramsey2012limitations, gorelick2017google}. Conversely, SAR sensors, like Sentinel-1, operate day and night and penetrate clouds and rain, offering dependable temporal coverage, but the backscatter data are difficult to interpret visually and often lack intuitive semantic cues about surface materials or flood-water extents \cite{manavalan2017sar, torres2012gmes}.

Existing data fusion strategies have made progress toward combining these complementary sensing modalities \cite{mahyoub2019fusing, fawakherji2025flood, fawakherji2024multi}. Yet, most current models treat SAR-to-optical translation and flood-water segmentation as separate tasks, trained with independent objectives \cite{wang2025mt_gan, pech2024segmentation}. Translation networks such as CycleGAN \cite{zhu2017unpaired} or pix2pix \cite{isola2017image} focus on reconstructing optical appearance but disregard hydrological semantics. Segmentation models such as U-Net \cite{ronneberger2015u} or Swin-Transformer \cite{liu2021swin} variants exploit SAR textures for water delineation but ignore the spectral realism of optical space. These isolated approaches overlook the physical coupling between reflectance reconstruction and flood-water region understanding, leading to either visually plausible but hydrologically inconsistent outputs or accurate masks that lack interpretability \cite{ma2022crossmodal}.

To bridge this gap, we propose FLoRA (Fusion-Latent for Optical Reconstruction and Flood-Water Area Segmentation), a cross-modal multi-task distillation framework that learns a shared latent representation from SAR and optical domains. Instead of fusing raw pixels or concatenated features, FLoRA introduces a fusion-latent space where SAR features are dynamically aligned with optical priors through multi-scale windowed cross-attention and Feature-wise Linear Modulation (FiLM) conditioning. A lightweight optical teacher, driven by RGB and NDVI cues, guides the SAR encoder toward perceptually consistent features, while gated residuals prevent over-correction from noisy priors.

Within this unified architecture, two decoders operate jointly: one reconstructs high-fidelity optical imagery, and the other segments flood-water regions. By enforcing structural and spectral consistency (via Charbonnier + SSIM + FFT losses) alongside hydrology-aware delineation (Dice + BCE + edge alignment), FLoRA achieves both visual realism and physical accuracy. A feature-distillation constraint further ensures that the fused SAR representation remains semantically anchored to the optical teacher’s manifold, enabling generalization across datasets and seasons.

Extensive experiments on DEEPFLOOD \cite{fawakherji2025deepflood}, SEN1FLOODS11 \cite{bonafilia2020sen1floods11} and SEN12MS \cite{schmitt2019sen12ms} demonstrate that FLoRA significantly outperforms prior fusion baselines in PSNR, SSIM, LPIPS, and Dice metrics. Qualitative analyses show that the model recovers fine spatial details of flooded and non-flooded regions while reconstructing realistic RGB structure. In summary, FLoRA establishes a new direction for cross-modal flood intelligence by unifying translation and segmentation through teacher-guided latent fusion, moving a step closer toward physically consistent, semantically interpretable disaster mapping from spaceborne observations. 

\textbf{Scope and Terminology Clarification:} It is important to distinguish between \textit{flood mapping} (typically a change detection task requiring pre- and post-event imagery) and \textit{flood-water body segmentation} (a state detection task). While FLoRA operates on single-temporal post-disaster SAR-optical pairs, we frame the task as \textit{flood-water segmentation}. In the context of rapid disaster response, pre-event baselines are not always immediately available or perfectly registered. Therefore, robustly segmenting all water bodies, including temporary inundation, from a single noisy SAR acquisition, guided by optical priors, serves as a critical first-order approximation for flood-water extent mapping. Throughout this work, the term ``flood-water segmentation'' refers to this robust extraction of water surfaces within flood-prone regions.

The three main contributions of FLoRA are as follows:
\begin{itemize}

\item \textbf{Unified Cross-Modal Framework:} FLoRA jointly performs SAR-to-optical translation and flood-water segmentation in one multi-task distillation architecture, replacing separate task-specific models while preserving both visual realism and hydrologic accuracy.

\item \textbf{Fusion-Latent Space with Optical Guidance:} A shared fusion-latent space aligns SAR features with optical priors through windowed cross-attention and FiLM conditioning. A lightweight optical teacher (RGB + NDVI) refines SAR representations using gated residuals to limit noise.

\item \textbf{Physically and Perceptually Consistent Training:} A composite loss, structural–spectral reconstruction (Charbonnier, SSIM, FFT) + hydrology-aware segmentation (Dice, BCE, edge terms) drives both spectral fidelity and hydrologic correctness, achieving state-of-the-art results on DEEPFLOOD, SEN1FLOODS11, and SEN12MS.

\end{itemize}

\begin{figure*}
{
  \centering
   \includegraphics[width=\linewidth]{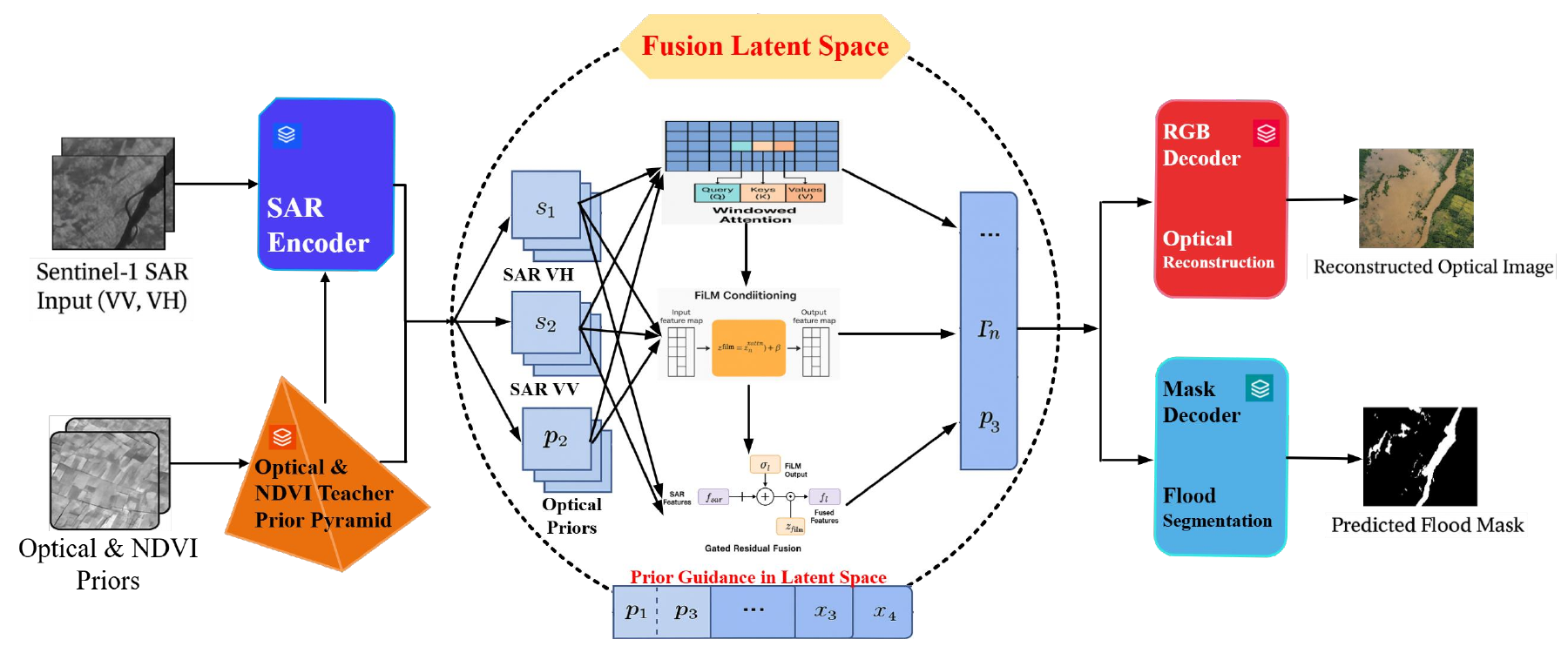}
    \caption{Overview of the proposed \textbf{FLoRA (Fusion-Latent for Optical Reconstruction and Flood-Water Area Segmentation)} framework.}
      \label{fig:flora_architecture}
}
\end{figure*}

\section{Related Work}
\label{sec:related}

Flood mapping methods \cite{mcglade2019global, hashemi2021flood} have evolved significantly from early physics-based thresholding and spectral indices to modern deep learning and cross-modal fusion frameworks. This section reviews the major directions of prior research, grouped into: (i) early flood-water mapping and single-modal deep learning,  (ii) cross-modal fusion  for joint reconstruction and segmentation, (iii) multi-task learning and student-teacher distillation, and (iv) need for unified cross-modal multi-task fusion.

\subsection{Early Flood-Water Mapping and Single-Modal Deep Learning}
Traditional flood-water mapping relied on statistical and spectral techniques such as the Normalized Difference Water Index (NDWI) and Modified NDWI (MNDWI) to delineate water bodies \cite{mcfeeters1996use}. While computationally simple and interpretable, these methods perform poorly in heterogeneous landscapes, mixed land cover, and shadowed regions \cite{ji2009analysis}.  

The introduction of deep learning revolutionized flood-water mapping by allowing neural networks to learn spatial hierarchies and contextual patterns directly from raw imagery \cite{zhu2017deep}. Architectures such as U-Net \cite{ronneberger2015u}, SegNet \cite{badrinarayanan2017segnet}, and ResU-Net \cite{zhang2018road} demonstrated substantial improvements in flood-water segmentation, particularly using Sentinel-1 SAR data. These convolutional networks exploit spatial context to distinguish flooded and non-flooded regions effectively. Recently, transformer-based models such as \cite {yang2023instance} and Swin-Transformer \cite{liu2021swin} have further improved performance by capturing long-range dependencies and global contextual information. However, these single-modal models remain constrained. Optical models fail in cloudy or nighttime conditions, while SAR-based models, though robust, struggle to provide visually or semantically interpretable results due to the absence of spectral cues.

\subsection{Cross-Modal Fusion and SAR-to-Optical Translation}
To address the limitations of single modalities, recent studies have explored fusion strategies that combine SAR and optical imagery \cite{salem2022inundated}. Early fusion approaches concatenated pixel-level \cite{kulkarni2020pixel} or feature-level representations \cite{yuan2018research}, while more advanced techniques applied attention mechanisms, gating networks, or adaptive weighting to merge features from both sensors \cite{fauvel2012advances}. Despite these improvements, many fusion methods still fail to fully exploit the complementary nature of the modalities, especially the spectral semantics provided by optical data.

With the rise of Generative Adversarial Networks (GANs), image-to-image translation frameworks such as CycleGAN \cite{zhu2017unpaired} and Pix2Pix \cite{isola2017image} introduced new possibilities for translating SAR images into optical space. These models treat SAR-to-optical translation as a form of cross-domain style transfer, improving visual realism and interpretability. However, they often neglect the hydrological semantics critical for flood-water analysis. As a result, their reconstructed optical outputs may appear visually convincing but fail to represent the true flood-water extent or water dynamics. Moreover, these translation models typically optimize only for image appearance, without integrating flood-water segmentation as a concurrent learning objective, which limits their generalization and practical utility in disaster scenarios \cite{meraner2020cloud, wang2025mt_gan}.

\subsection{Multi-Task Learning and Teacher-Student Distillation}
Multi-task learning (MTL) has emerged as a powerful paradigm for improving model generalization by jointly optimizing multiple related objectives. In the context of remote sensing, MTL frameworks \cite{shen2022coupling} have demonstrated that shared latent representations between tasks such as image reconstruction and segmentation can improve both accuracy and feature consistency. By learning from complementary objectives, models can better capture both the geometric and semantic structure of the scene.

Teacher-student frameworks extend this idea further by using feature distillation to guide learning across modalities. In such settings, a teacher model, typically trained on optical imagery, provides high-level guidance to a SAR-based student network \cite{fan2022novel}. This supervision enables the SAR model to capture more semantically meaningful representations that align with optical visual cues \cite{song2024efficient}. While effective, most of these approaches address either the translation or segmentation problem in isolation, rather than optimizing both jointly, thus failing to capture the coupled nature of visual reconstruction and hydrological interpretation.

\subsection{The Need for Unified Cross-Modal Multi-Task Fusion}
Despite the progress in both flood-water segmentation and SAR-to-optical translation, existing models largely treat these problems separately. SAR-based segmentation approaches, from early thresholding \cite{otsu1975threshold} to modern architectures such as U-Net \cite{ronneberger2015u} and Swin-Transformer \cite{liu2021swin}, are effective at identifying flooded areas but rely solely on the statistical and textural properties of backscatter. This makes it difficult to distinguish between visually similar surfaces such as flooded vegetation, urban reflections, or saturated soil. Conversely, optical reconstruction models generate visually plausible imagery but often fail to preserve hydrological consistency, producing outputs that do not align with true flood-water dynamics.  

These challenges emphasize the need for a unified framework that integrates both modalities and tasks within a single learning paradigm. Such a cross-modal multi-task approach should align optical priors with SAR features, ensuring that visually realistic reconstructions correspond to physically accurate flood-water boundaries. The combination of spectral information from optical imagery and the geometric and temporal consistency of SAR can significantly improve interpretability and reliability.  

Our proposed framework, \textbf{FLoRA}, directly addresses this gap by introducing a cross-modal multi-task learning strategy that fuses SAR and optical representations within a shared latent space. Through joint optimization of optical reconstruction and flood-water segmentation, FLoRA achieves both high visual fidelity and hydrologically consistent flood-water mapping.

\section{Methodology}
\label{sec:methodology}

Flood-water mapping from spaceborne data poses two major challenges: the spectral ambiguity of SAR backscatter and the lack of cloud-free optical imagery during disasters. 
The proposed \textbf{FLoRA (Fusion-Latent for Optical Reconstruction and Flood-Water Area Segmentation)} framework addresses these issues through a \emph{cross-modal multi-task distillation network} that unifies SAR and optical feature spaces. 
By constructing a \textit{fusion-latent space} that captures both the geometric properties of SAR and the semantic richness of optical data, FLoRA jointly reconstructs realistic optical imagery and segments flood-water regions in a physically consistent manner.

\textit{Figure 1 illustrates the overall data flow of FLoRA. Given SAR input and optional optical priors, features are first extracted by modality-specific encoders. These features are aligned within a fusion-latent space using windowed cross-attention, FiLM conditioning, and gated residual fusion. The fused representation is then shared across two gradient-decoupled decoders for optical reconstruction and floodwater segmentation.}

Let $S\!\in\!\mathbb{R}^{2\times H\times W}$ denote the Sentinel-1 SAR input (VV, VH) and 
$P\!\in\!\mathbb{R}^{C_p\times H\times W}$ represent the optical prior, such as Sentinel-2 RGB or NDVI. 
The network learns a mapping 
\begin{equation}
F(S, P) \longrightarrow (\widehat{Y}, \widehat{M}),
\label{eq:overall_mapping}
\end{equation}
where $\widehat{Y}$ is the reconstructed optical image and $\widehat{M}$ is the predicted flood-water mask.  

\noindent\textbf{Pipeline summary: } 
\begin{equation}
(S, P)\rightarrow Encoders \rightarrow FusionLatent \rightarrow ({DEC}_{RGB},{DEC}_{Mask})
\label{eq:pipeline}
\end{equation}

The core of FLoRA is the \emph{fusion-latent space}, which aligns features from both modalities through multi-scale windowed cross-attention, Feature-wise Linear Modulation (FiLM) and gated residual fusion. To stabilize optimization, FLoRA introduces gradient-decoupled dual decoders and a teacher–student distillation that transfers optical semantics into SAR representations. 

\subsection{Encoders and Teacher Paths}

\paragraph{SAR Encoder.}
A U-Net-like encoder progressively extracts structural features from SAR backscatter:
\begin{equation}
\{ \mathbf{s}_1,\mathbf{s}_2,\mathbf{s}_3,\mathbf{x}_m \}
  = \mathrm{Enc}_{\text{SAR}}(S),
\label{eq:sar_encoder}
\end{equation}
where
$\mathbf{s}_1\!\in\!\mathbb{R}^{B\times \frac{H}{2}\times \frac{W}{2}}$,
$\mathbf{s}_2\!\in\!\mathbb{R}^{2B\times \frac{H}{4}\times \frac{W}{4}}$,
$\mathbf{s}_3\!\in\!\mathbb{R}^{4B\times \frac{H}{8}\times \frac{W}{8}}$,
and $\mathbf{x}_m\!\in\!\mathbb{R}^{8B\times \frac{H}{8}\times \frac{W}{8}}$.

\textit{These encoder outputs provide multi-scale features that serve as inputs to the fusion-latent alignment module.}

\paragraph{Optical Teacher Pyramid.}
A lightweight optical teacher extracts hierarchical features from $P$:
\begin{equation}
\mathcal{P} = \{ \mathbf{p}_1,\mathbf{p}_2,\mathbf{p}_3,\mathbf{p}_4 \}
= \mathrm{Teach}(P),
\label{eq:teacher}
\end{equation}
where $\mathbf{p}_\ell\!\in\!\mathbb{R}^{(2^{\ell-1}B)\times \frac{H}{2^\ell}\times \frac{W}{2^\ell}}$. 
When optical priors are unavailable, a SAR-driven prior predictor $\mathrm{Prior}(S)$ provides surrogate guidance. 
This design ensures the same fusion process can be used in both training and inference.

The teacher is implemented as a lightweight convolutional feature pyramid (Conv--GroupNorm--SiLU blocks with progressive pooling) as seen in Figure 2, that provides multi-scale guidance features rather than final predictions. These priors guide SAR feature alignment inside the fusion-latent space through cross-attention and FiLM conditioning.

When optical priors are unavailable, a SAR-driven prior predictor $\mathrm{Prior}(S)$ generates surrogate prior features with the same pyramid structure:
\begin{equation}
\{ \mathbf{p}_1,\mathbf{p}_2,\mathbf{p}_3,\mathbf{p}_4 \}
= \mathrm{Prior}(S),
\end{equation}
allowing the same fusion pathway to be used during both training (optical-guided) and inference (SAR-only) without architectural changes.

\begin{figure}[t]
{
  \centering
   \includegraphics[width=\linewidth]{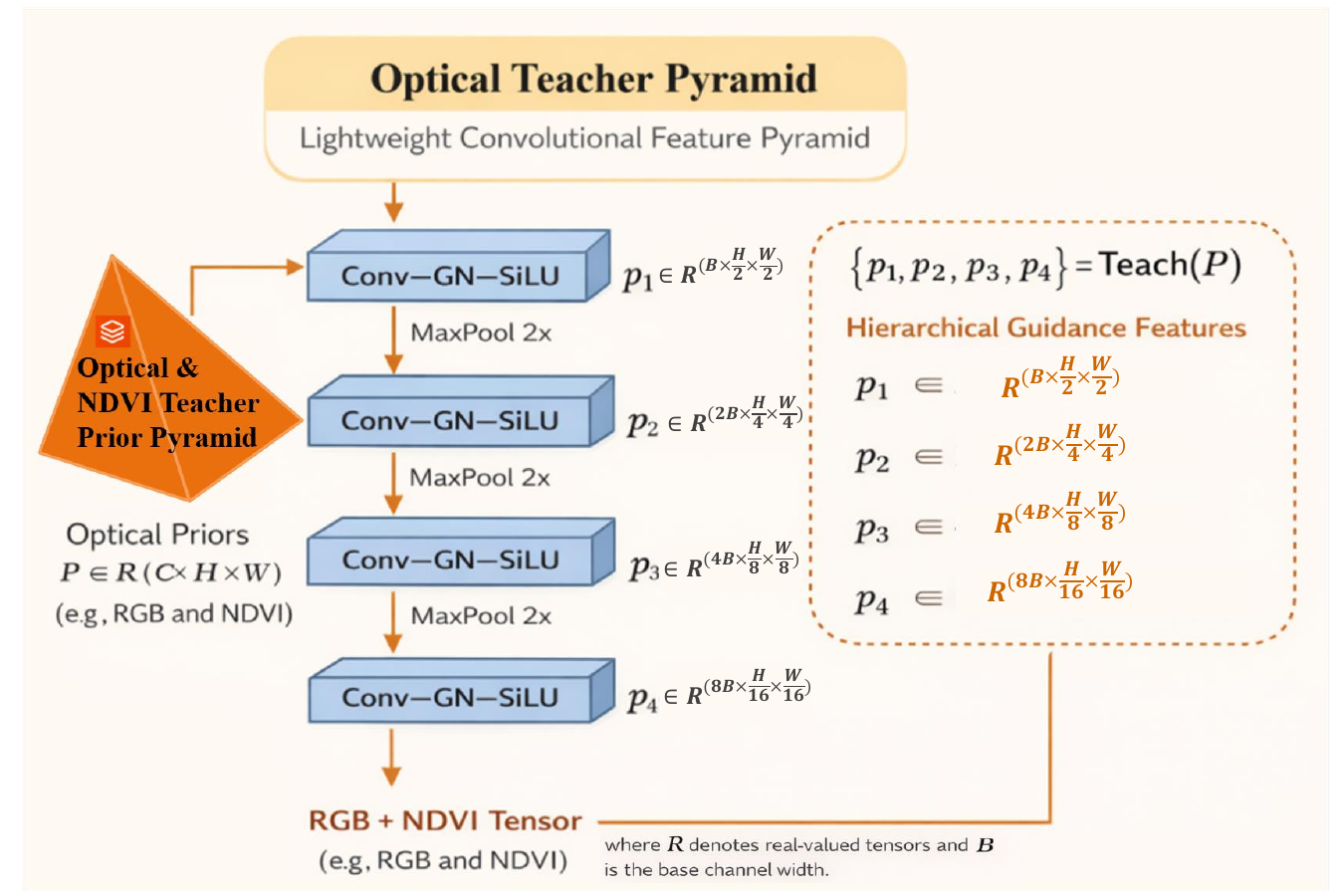}
    \caption{Architecture of the lightweight optical teacher pyramid. Given optical priors $P$, the teacher extracts hierarchical guidance features $\{\mathbf{p}_1,\mathbf{p}_2,\mathbf{p}_3,\mathbf{p}_4\}$ at multiple scales, which are used to guide SAR feature fusion in the latent space.}
      \label{fig:opt_teacher}
}
\end{figure}

\subsection{Fusion-Latent Space: Cross-Modal Alignment}
\label{sec:fusion_latent}

SAR and optical modalities differ significantly in signal distribution: SAR encodes geometric and backscatter structure, while optical data contain semantic and spectral cues.  Direct concatenation of these modalities can lead to feature-space mismatches and noisy gradients. 
To address this, FLoRA introduces a dedicated \textbf{fusion-latent space} that harmonizes both modalities at multiple scales using three complementary operations: 
(i) localized \textit{windowed cross-attention}, 
(ii) adaptive \textit{FiLM conditioning}, and 
(iii) confidence-aware \textit{gated residual fusion}. 
Together, these mechanisms ensure that the SAR encoder benefits from optical priors without losing its inherent geometric integrity.

\paragraph{Windowed Cross-Attention}

At each feature scale $\ell$, the SAR encoder produces intermediate feature maps $\mathbf{f}^{\text{sar}}_\ell \in \mathbb{R}^{C\times H_\ell\times W_\ell}$, 
while the optical teacher (or prior predictor) produces $\mathbf{f}^{\text{opt}}_\ell \in \mathbb{R}^{C\times H_\ell\times W_\ell}$ of the same spatial resolution. 
Windowed cross-attention is applied using non-overlapping local windows of size $w=8$ and $h=4$ attention heads at all pyramid levels.
Given query features from the SAR encoder and key/value features from the optical prior pyramid, attention is computed independently within each $8\times8$ spatial window to improve computational efficiency and preserve local spatial structure. The window size is chosen empirically as a balance between local contextual modeling and computational cost. Unlike Swin Transformer, our implementation does not employ relative positional bias. Spatial consistency is instead preserved through the convolutional encoder hierarchy and multi-scale fusion design.
These maps are first linearly projected into \emph{query}, \emph{key}, and \emph{value} tensors:
\begin{equation}
\mathbf{Q}=\mathbf{W}^Q\!\ast\!\mathbf{f}^{\text{sar}}_\ell,\quad
\mathbf{K}=\mathbf{W}^K\!\ast\!\mathbf{f}^{\text{opt}}_\ell,\quad
\mathbf{V}=\mathbf{W}^V\!\ast\!\mathbf{f}^{\text{opt}}_\ell,
\label{eq:qkv}
\end{equation}
where $\mathbf{W}^Q$, $\mathbf{W}^K$, and $\mathbf{W}^V$ are $1{\times}1$ convolutional projections and $\ast$ denotes convolution. 
$\mathbf{Q}$ encodes what the SAR modality seeks to learn, while $\mathbf{K}$ and $\mathbf{V}$ represent the optical domain’s reference information.

To capture local spatial correspondences efficiently, the feature maps are partitioned into non-overlapping windows of size $w{\times}w$. 
Within each window $i$, attention is computed as
\begin{equation}
\mathrm{Attn}(\mathbf{Q}_i,\mathbf{K}_i,\mathbf{V}_i)
=\mathrm{softmax}\!\left(\frac{\mathbf{Q}_i \mathbf{K}_i^\top}{\sqrt{d_h}}\right)\mathbf{V}_i,
\label{eq:attention}
\end{equation}
where $d_h$ is the feature dimension per attention head. 
This localized design restricts correspondence learning to spatially relevant neighborhoods, critical for terrain-dependent flood-water patterns, while reducing computational cost. 
All attended patches are then aggregated and projected back to full resolution:
\begin{equation}
\mathbf{z}^{\text{xattn}}_\ell 
  = \mathbf{W}^O \!\ast\!
  \mathrm{Fold}\!\big(\{\mathrm{Attn}(\mathbf{Q}_i,\mathbf{K}_i,\mathbf{V}_i)\}_i\big),
\label{eq:xattn_output}
\end{equation}
where $\mathbf{W}^O$ is an output projection and $\mathrm{Fold}(\cdot)$ reassembles the windowed outputs. 
The resulting tensor $\mathbf{z}^{\text{xattn}}_\ell$ thus embeds each SAR feature with context-aware optical semantics, creating a locally aligned latent representation.

\paragraph{FiLM Conditioning}

Even after attention alignment, SAR and optical features differ statistically due to distinct imaging mechanisms. 
To adaptively calibrate SAR features using optical priors, FLoRA employs \textit{Feature-wise Linear Modulation (FiLM)}. 
This mechanism learns channel-wise scaling and shifting parameters $(\gamma_\ell,\beta_\ell)$ from the optical features:
\begin{equation}
[\gamma_\ell,\beta_\ell]=\Phi(\mathbf{f}^{\text{opt}}_\ell), \qquad
\mathbf{z}^{\text{film}}_\ell
  = \mathbf{z}^{\text{xattn}}_\ell \odot (1+\tanh(\gamma_\ell)) + \beta_\ell.
\label{eq:film}
\end{equation}
Here, $\Phi(\cdot)$ is a small convolutional subnetwork that predicts modulation coefficients, 
$\odot$ denotes element-wise multiplication, and $\tanh(\cdot)$ stabilizes the scaling factor around unity. 
Intuitively, $\gamma_\ell$ controls the amplitude of feature channels based on optical cues (e.g., vegetation, water, or urban patterns), while $\beta_\ell$ introduces additive corrections. 
FiLM therefore acts as a learned normalization that brings SAR activations closer to their optical counterparts without overwriting physical structures such as backscatter boundaries.

\paragraph{Gated Residual Fusion}

Although cross-attention and FiLM help align modalities, over-reliance on optical priors can distort physically valid SAR cues (e.g., flooded areas under cloud cover). 
To address this, FLoRA introduces a confidence-aware gate that determines how strongly optical corrections should influence the fused representation:
\begin{equation}
\mathbf{f}_\ell
  = \mathbf{f}^{\text{sar}}_\ell 
  + \mathbf{g}_\ell \odot 
  \big(\mathbf{z}^{\text{film}}_\ell - \mathbf{f}^{\text{sar}}_\ell\big),
\quad \mathbf{g}_\ell=\sigma(\mathbf{W}^g\!\ast\!\mathbf{f}^{\text{sar}}_\ell),
\label{eq:gated}
\end{equation}
where $\sigma(\cdot)$ is the sigmoid activation producing a gating tensor $\mathbf{g}_\ell \in [0,1]^{C\times H_\ell\times W_\ell}$ that reflects the model’s confidence in optical information. 
If the SAR signal is reliable (e.g., low speckle and clear surface backscatter), the gate suppresses corrections; 
when the signal is ambiguous (e.g., shadow or flooded vegetation), the gate allows stronger optical influence. 
The final fused feature $\mathbf{f}_\ell$ thus adaptively integrates both modalities, preserving SAR’s geometric fidelity while inheriting optical semantics.

\paragraph{Multi-Scale Fusion Summary}

This process is applied hierarchically across four pyramid levels, yielding the multi-scale latent set
\begin{equation}
\mathcal{F}=\{\mathbf{f}_1,\mathbf{f}_2,\mathbf{f}_3,\mathbf{f}_4\},
\end{equation}
where shallower levels ($\mathbf{f}_1,\mathbf{f}_2$) capture fine-grained texture alignment and deeper levels ($\mathbf{f}_3,\mathbf{f}_4$) encode contextual hydrological consistency. 
The resulting \textit{fusion-latent space} provides a compact, semantically aligned representation that supports both optical reconstruction and flood-water segmentation tasks downstream. The resulting fused features are passed to the multi-task decoders.

\subsection{Multi-Task Learning with Gradient Decoupling}
\label{sec:grad_decouple}

The two objectives in FLoRA-optical reconstruction and flood-water segmentation exhibit inherently different optimization tendencies. While reconstruction prioritizes spectral and textural fidelity, segmentation emphasizes geometric and boundary accuracy. 
Naïvely training both heads from the same latent features often causes \textit{gradient conflict}, where competing updates destabilize the shared encoder. 

To mitigate this, FLoRA introduces \textbf{gradient decoupling}, which selectively blocks gradient flow between decoders. 
Let $\mathcal{F}$ denote the fused multi-scale features from the fusion-latent space, and $\operatorname{sg}(\cdot)$ the stop-gradient operator that halts backpropagation. 
The features used by each decoder are defined as:
\begin{equation}
\begin{aligned}
\mathcal{F}^{\text{seg}} &=
\begin{cases}
\operatorname{sg}(\mathcal{F}), & \text{if } \texttt{seg\_from\_rgb=True},\\
\mathcal{F}, & \text{otherwise,}
\end{cases}\\[3pt]
\mathcal{F}^{\text{rgb}} &=
\begin{cases}
\operatorname{sg}(\mathcal{F}), & \text{if } \texttt{rgb\_from\_seg=True},\\
\mathcal{F}, & \text{otherwise.}
\end{cases}
\end{aligned}
\label{eq:gradstop}
\end{equation}

By default, FLoRA sets \texttt{seg\_from\_rgb=True}, ensuring segmentation gradients are blocked from the shared encoder. 
This allows the reconstruction branch to govern cross-modal alignment (learning optical semantics), while the segmentation head focuses on spatial delineation without perturbing the shared representation. 
Conversely, setting \texttt{rgb\_from\_seg=True} can prioritize hydrologic structure in joint learning scenarios. 

This simple gating mechanism effectively disentangles task supervision while preserving semantic coupling through shared features.

To avoid competing optimization objectives between reconstruction and segmentation, gradient decoupling is introduced before the dual decoders.

\subsection{Dual Decoder Design}
\label{sec:dual_decoder}

The fusion-latent representation $\mathcal{F}$ feeds two specialized decoders that learn complementary objectives under the decoupling regime. 

\paragraph{RGB Decoder (Optical Reconstruction).}
The RGB decoder $\mathrm{Dec}_{\text{RGB}}$ employs progressive upsampling with skip connections to recover fine-grained spatial detail and spectral consistency:
\begin{equation}
\widehat{Y} = \sigma(\mathrm{Dec}_{\text{RGB}}(\mathcal{F}^{\text{rgb}})),
\label{eq:rgbdecoder}
\end{equation}
where $\widehat{Y}\!\in\!\mathbb{R}^{3\times H\times W}$ is the reconstructed optical image and $\sigma(\cdot)$ is a sigmoid activation. 
This branch is trained using structural and spectral consistency losses (Charbonnier, SSIM, FFT, edge), promoting realistic reflectance reconstruction aligned with optical priors.

\paragraph{Mask Decoder (Flood-Water Segmentation).}
The mask decoder $\mathrm{Dec}_{\text{Mask}}$ mirrors the RGB decoder’s structure but outputs a single-channel flood-water probability map:
\begin{equation}
\widehat{M} = \sigma(\mathrm{Dec}_{\text{Mask}}(\mathcal{F}^{\text{seg}})).
\label{eq:maskdecoder}
\end{equation}
It is supervised via Dice, BCE, and hydrology-aware edge losses, focusing on precise boundary delineation of flooded areas. 

Together, the two decoders act as cooperative learners: the reconstruction path enforces spectral realism and contextual awareness, while the segmentation path constrains spatial and hydrologic structure. 
Their joint optimization within the decoupled regime ensures that shared features remain semantically rich yet physically interpretable, enabling FLoRA to generate visually consistent and hydrologically reliable outputs.

\subsection{Loss Formulation}

\paragraph{Reconstruction Losses}
To ensure optical realism and spatial coherence, we use:
\begin{align}
\mathcal{L}_{\text{charb}} &= \frac{1}{HW}\sum\sqrt{(\widehat{Y}-Y)^2+\epsilon}, \label{eq:charb}\\
\mathcal{L}_{\text{ssim}} &= 1-\mathrm{SSIM}_{3\times3}(\widehat{Y},Y), \label{eq:ssim}\\
\mathcal{L}_{\text{fft}}  &= \big\| \log|\mathcal{F}(\bar{Y})|-\log|\mathcal{F}(\bar{\widehat{Y}})| \big\|_1, \label{eq:fft}\\
\mathcal{L}_{\text{edge}} &= \big\| \|\nabla\bar{\widehat{Y}}\|_2 - \|\nabla\bar{Y}\|_2 \big\|_1. \label{eq:edge}
\end{align}
The reconstruction loss is
\begin{equation}
\mathcal{L}_{\text{RGB}} =
\lambda_c\mathcal{L}_{\text{charb}}+
\lambda_s\mathcal{L}_{\text{ssim}}+
\lambda_f\mathcal{L}_{\text{fft}}+
\lambda_e\mathcal{L}_{\text{edge}}.
\label{eq:lrgb}
\end{equation}

\paragraph{Segmentation Losses}
Flood-water delineation accuracy is encouraged by combining overlap, pixel, and boundary constraints:
\begin{align}
\mathcal{L}_{\text{dice}} &= 1 - \frac{2\langle \widehat{M},M\rangle}{\|\widehat{M}\|_1 + \|M\|_1 + \epsilon}, \label{eq:dice}\\
\mathcal{L}_{\text{bce}} &= -\frac{1}{HW}\sum \big[M\log\widehat{M} + (1-M)\log(1-\widehat{M})\big], \label{eq:bce}\\
\mathcal{L}_{\text{hydro}} &= \frac{1}{|\mathcal{B}|}\sum_{(u,v)\in\mathcal{B}}\!\Big(1-\cos\angle(\nabla\bar{\widehat{Y}}(u,v),\nabla\bar{S}(u,v))\Big). \label{eq:hydro}
\end{align}
The combined segmentation objective is:
\begin{equation}
\mathcal{L}_{\text{SEG}} =
\mu_d\mathcal{L}_{\text{dice}}+
\mu_b\mathcal{L}_{\text{bce}}+
\mu_h\mathcal{L}_{\text{hydro}}.
\label{eq:lseg}
\end{equation}

\paragraph{Feature Distillation}
A teacher–student loss enforces alignment of fused SAR features with the optical manifold:
\begin{equation}
\mathcal{L}_{\text{distill}} = 
\sum_{\ell=1}^{4}\|\mathbf{f}_\ell - \operatorname{sg}(\mathbf{p}_\ell)\|_1.
\label{eq:ldistill}
\end{equation}

\paragraph{Overall Objective}
The total training loss is the weighted sum:
\begin{equation}
\mathcal{L}_{\text{total}} =
\mathcal{L}_{\text{RGB}} +
\mathcal{L}_{\text{SEG}} +
\eta\,\mathcal{L}_{\text{distill}}.
\label{eq:ltotal}
\end{equation}

In summary, FLoRA introduces a cross-modal fusion-latent framework that bridges the spectral–structural features between SAR and optical imagery.  By coupling multi-scale attention, FiLM conditioning, and gradient-decoupled dual decoding, the model jointly reconstructs semantically accurate optical reflectance and hydrologically consistent flood-water masks. Teacher-guided feature distillation anchors the learning process to physically meaningful priors, enabling FLoRA to produce semantically faithful and physically coherent flood intelligence even under SAR-only conditions.

\section{Experiments and Results}
\label{sec:experiments}

\subsection{Datasets}
The proposed \textbf{FLoRA} framework was evaluated on two benchmark datasets: DEEPFLOOD \cite{fawakherji2025deepflood}, SEN1FLOODS11  \cite{bonafilia2020sen1floods11}, and SEN12MS \cite{schmitt2019sen12ms}, covering diverse flood events and geographic conditions.

\textbf{DEEPFLOOD} \cite{fawakherji2025deepflood} provides Sentinel-1, Sentinel-2, UAV, DEM, and water-index layers across multiple North Carolina flood events. We use its SAR\_VV, SAR\_VH, optical, and water-index data for evaluating reconstruction and segmentation.

\textbf{SEN1FLOODS11}~\cite{bonafilia2020sen1floods11} provides globally distributed Sentinel-1 (VV/VH) and Sentinel-2 pairs with flood annotations. It contains approximately 4{,}446 training and 1{,}050 test tiles at $512\times512$ resolution. We use the \textit{hand-labeled subset} for supervised SAR-to-optical translation and flood-water segmentation experiments.

\textbf{SEN12MS}~\cite{schmitt2019sen12ms} consists of over 180{,}000 co-registered Sentinel-1/2 patches of size $256\times256$ across four seasonal subsets. In this work, we use only the \textbf{winter subset} (about 43{,}000 patches), which offers stronger SAR--optical contrast and supports more robust cross-modal learning.

\subsection{Training and Experimental Setup}

All images were resampled to 10\,m resolution, normalized to $[0,1]$, and tiled into $256\times256$ patches. Models were trained for 1000 epochs using \texttt{Python 3.10} and \texttt{PyTorch>1.10} with mixed-precision on an NVIDIA RTX~4090 GPU and NVIDIA A4000 GPU.
The AdamW optimizer was used with an initial learning rate of $1\times10^{-4}$ and cosine decay over 200 epochs. Unless otherwise stated, loss weighting followed a ratio of $\mathcal{L}_{\text{RGB}} : \mathcal{L}_{\text{Seg}} = 1 : 0.5$.
All baseline models were re-trained using the same training/validation/test splits, input resolution, and preprocessing pipeline.
For segmentation comparisons, all models receive identical Sentinel-1 SAR inputs (VV, VH) without additional optical priors.
For SAR-to-optical translation baselines (MT\_GAN \cite{wang2025mtgan}, Nice\_GAN \cite{chen2020reusing}, TS2O \cite{zhang2025ts2o}, CycleGAN \cite{zhu2017unpaired}, and pix2pix \cite{isola2017image}), training follows their original loss formulations while maintaining a unified data protocol for fair comparison.

\textbf{Evaluation Metrics:} We evaluate both reconstruction and segmentation performance using standard quantitative metrics.
For optical reconstruction, we report PSNR, SSIM, and LPIPS to jointly measure pixel-level fidelity, structural similarity, and perceptual quality.
For flood-water segmentation, multiple complementary metrics are used, including IoU, Dice, Precision, Recall, and F1 score.
IoU and Dice quantify region overlap, while Precision and Recall characterize false-positive and false-negative behavior, respectively.
The F1 score is additionally reported as the harmonic mean of Precision and Recall to provide a balanced assessment under class imbalance.
Reporting this set of metrics ensures consistent and comprehensive evaluation across datasets and allows fair comparison with prior remote sensing studies.
All metrics were averaged over 1{,}000 validation tiles per dataset.

\subsection{Quantitative Results}
FLoRA is compared against both classical and recent SAR-to-optical translation methods, including U-Net, MT\_GAN \cite{wang2025mtgan}, Nice\_GAN \cite{chen2020reusing}, TS2O \cite{zhang2025ts2o}, CycleGAN \cite{zhu2017unpaired}, TransUNet \cite{zheng2025based}, and pix2pix \cite{isola2017image}. 
For flood-water segmentation, comparisons include U-Net \cite{ronneberger2015u}, MSResNet \cite{adhikari2025msresnet}, TransUNet \cite{zheng2025based}, and GAN-based fusion baselines. Tables I–III report optical reconstruction performance, while Tables IV–VI present flood-water segmentation results.

\begin{table}[t]
\centering
\caption{Optical reconstruction performance on SEN1FLOODS11 \cite{bonafilia2020sen1floods11} dataset.}
\begin{tabular}{lccc}
\toprule
\textbf{Method} & \textbf{PSNR$\uparrow$} & \textbf{SSIM$\uparrow$} & \textbf{LPIPS$\downarrow$} \\
\midrule
U-Net \cite{ronneberger2015u} & 22.81 & 0.6413 & 0.701 \\
MT\_GAN \cite{wang2025mtgan} & 23.28 & 0.6648 & 0.695 \\
Nice\_GAN \cite{chen2020reusing} & 23.59 & 0.6792 & 0.686 \\
TS2O \cite{zhang2025ts2o} & 24.01 & 0.6912 & 0.681 \\
CycleGAN \cite{zhu2017unpaired} & 24.12 & 0.7045 & 0.674 \\
TransUNet \cite{zheng2025based} & 25.38 & 0.7216 & 0.662 \\
pix2pix  \cite{isola2017image} & 29.56 & 0.7428 & 0.587 \\
\textbf{FLoRA (ours)} & \textbf{33.61} & \textbf{0.8124} & \textbf{0.4533} \\
\bottomrule
\end{tabular}
\label{tab:quantitative_recon}
\end{table}

\begin{table}[t]
\centering
\caption{Optical reconstruction performance on SEN12MS \cite{schmitt2019sen12ms} dataset.}
\begin{tabular}{lccc}
\toprule
\textbf{Method} & \textbf{PSNR$\uparrow$} & \textbf{SSIM$\uparrow$} & \textbf{LPIPS$\downarrow$} \\
\midrule
U-Net \cite{ronneberger2015u} & 21.64 & 0.5914 & 0.693 \\
MT\_GAN \cite{wang2025mtgan} & 21.82 & 0.6012 & 0.689 \\
Nice\_GAN \cite{chen2020reusing} & 22.36 & 0.6112 & 0.681 \\
TS2O \cite{zhang2025ts2o} & 22.92 & 0.6214 & 0.674 \\
CycleGAN \cite{zhu2017unpaired} & 23.52 & 0.6328 & 0.663 \\
TransUNet \cite{zheng2025based} & 25.71 & 0.6742 & 0.651 \\
pix2pix  \cite{isola2017image} & 26.88 & 0.7011 & 0.576 \\
\textbf{FLoRA (ours)} & \textbf{28.31} & \textbf{0.7530} & \textbf{0.4273} \\
\bottomrule
\end{tabular}
\label{tab:quantitative_sen12ms}
\end{table}

\begin{table}[t]
\centering
\caption{Optical reconstruction performance on the DEEPFLOOD \cite{fawakherji2025deepflood} dataset.}
\begin{tabular}{lccc}
\toprule
\textbf{Method} & \textbf{PSNR$\uparrow$} & \textbf{SSIM$\uparrow$} & \textbf{LPIPS$\downarrow$} \\
\midrule
U-Net \cite{ronneberger2015u}        & 22.18 & 0.6037 & 0.685 \\
MT\_GAN \cite{wang2025mtgan} & 22.56 & 0.6118 & 0.673 \\
Nice\_GAN \cite{chen2020reusing} & 23.16 & 0.6325 & 0.664 \\
TS2O \cite{zhang2025ts2o} & 23.75 & 0.6431 & 0.652 \\
CycleGAN \cite{zhu2017unpaired}     & 24.05 & 0.6512 & 0.648 \\
TransUNet \cite{zheng2025based}    & 26.14 & 0.6895 & 0.629 \\
pix2pix \cite{isola2017image}       & 27.92 & 0.7243 & 0.552 \\
\textbf{FLoRA (ours)}               & \textbf{28.47} & \textbf{0.7416} & \textbf{0.4819} \\
\bottomrule
\end{tabular}
\label{tab:quantitative_deepflood}
\end{table}

For GAN-based translation baselines, segmentation masks are obtained by attaching the same segmentation decoder used in FLoRA to ensure consistent comparison.

\begin{table}[t]
\centering
\caption{Segmentation performance on SEN1FLOODS11 \cite{bonafilia2020sen1floods11} dataset.}
\begin{tabular}{lccccc}
\toprule
\textbf{Method} & \textbf{IoU$\uparrow$} & \textbf{Dice$\uparrow$} & \textbf{Precision$\uparrow$} & \textbf{Recall$\uparrow$} & \textbf{F1$\uparrow$} \\
\midrule
U-Net \cite{ronneberger2015u}  & 0.57 & 0.58 & 0.55 & 0.60 & 0.57 \\
MSResNet \cite{adhikari2025msresnet} & 0.59 & 0.61 & 0.65 & 0.66 & 0.62 \\
TransUNet \cite{zheng2025based} & 0.61 & 0.60 & 0.68 & 0.73 & 0.70 \\
CycleGAN \cite{zhu2017unpaired} & 0.63 & 0.62 & 0.70 & 0.75 & 0.72 \\
pix2pix \cite{isola2017image} & 0.66 & 0.64 & 0.72 & 0.77 & 0.74 \\
\textbf{FLoRA (ours)} & \textbf{0.71} & \textbf{0.68} & \textbf{0.77} & \textbf{0.79} & \textbf{0.78} \\
\bottomrule
\end{tabular}
\label{tab:seg_sen1floods11}
\end{table}

\begin{table}[t]
\centering
\caption{Segmentation performance on SEN12MS \cite{schmitt2019sen12ms} dataset.}
\begin{tabular}{lccccc}
\toprule
\textbf{Method} & \textbf{IoU$\uparrow$} & \textbf{Dice$\uparrow$} & \textbf{Precision$\uparrow$} & \textbf{Recall$\uparrow$} & \textbf{F1$\uparrow$} \\
\midrule
U-Net \cite{ronneberger2015u}  & 0.49 & 0.53 & 0.56 & 0.51 & 0.53 \\
MSResNet \cite{adhikari2025msresnet} & 0.51 & 0.56 & 0.57 & 0.59 & 0.56 \\
TransUNet \cite{zheng2025based} & 0.53 & 0.62 & 0.59 & 0.64 & 0.61 \\
CycleGAN \cite{zhu2017unpaired} & 0.55 & 0.63 & 0.61 & 0.66 & 0.63 \\
pix2pix \cite{isola2017image} & 0.58 & 0.65 & 0.63 & 0.68 & 0.65 \\
\textbf{FLoRA (ours)} & \textbf{0.63} & \textbf{0.69} & \textbf{0.68} & \textbf{0.71} & \textbf{0.69} \\
\bottomrule
\end{tabular}
\label{tab:seg_sen12ms}
\end{table}

\begin{table}[t]
\centering
\caption{Segmentation performance on the DEEPFLOOD \cite{fawakherji2025deepflood} dataset.}
\begin{tabular}{lccccc}
\toprule
\textbf{Method} & \textbf{IoU$\uparrow$} & \textbf{Dice$\uparrow$} & \textbf{Precision$\uparrow$} & \textbf{Recall$\uparrow$} & \textbf{F1$\uparrow$} \\
\midrule
U-Net \cite{ronneberger2015u}  & 0.51 & 0.55 & 0.58 & 0.54 & 0.56 \\
MSResNet \cite{adhikari2025msresnet} & 0.52 & 0.57 & 0.59 & 0.63 & 0.61 \\
TransUNet \cite{zheng2025based} & 0.55 & 0.61 & 0.62 & 0.66 & 0.64 \\
CycleGAN \cite{zhu2017unpaired} & 0.57 & 0.63 & 0.64 & 0.68 & 0.66 \\
pix2pix \cite{isola2017image} & 0.60 & 0.66 & 0.66 & 0.70 & 0.68 \\
\textbf{FLoRA (ours)} & \textbf{0.64} & \textbf{0.71} & \textbf{0.71} & \textbf{0.74} & \textbf{0.72} \\
\bottomrule
\end{tabular}
\label{tab:seg_deepflood}
\end{table}

FLoRA yields an average gain over all datasets of $+2.6$\,dB in PSNR and $+8$\% in IoU over the strongest baseline (TransUNet). The improvement in LPIPS indicates perceptually more realistic reconstructions, while higher Dice scores demonstrate better delineation of flood-water boundaries.

\subsection{Qualitative Analysis}

Figures 3, 4, and 5 present representative qualitative comparisons between FLoRA and competing methods across the SEN1FLOODS11 \cite{bonafilia2020sen1floods11}, SEN12MS \cite{schmitt2019sen12ms}, and DEEPFLOOD \cite{fawakherji2025deepflood} datasets. The results include both optical reconstruction and flood-water segmentation outputs.

To improve interpretability, each figure includes annotated regions (red and green boxes) highlighting challenging scenarios such as shoreline boundaries and regions with ambiguous SAR backscatter. Arrows and descriptive labels are used to explicitly indicate these regions. Ground-truth flood-water masks are displayed alongside predicted masks, which are thresholded to binary, to enable direct and consistent visual comparison.

From these examples, FLoRA produces visually sharper and more spatially consistent reconstructions relative to baseline methods. In the annotated regions, competing models often exhibit blurred boundaries or false detections, while FLoRA generally maintains boundary continuity and reduces false positives, particularly in regions with low backscatter or complex land cover. These qualitative observations align with quantitative improvements in LPIPS and Dice metrics reported in Tables I-VI.

\begin{figure*}[t]
  \centering
  \includegraphics[width=\linewidth]{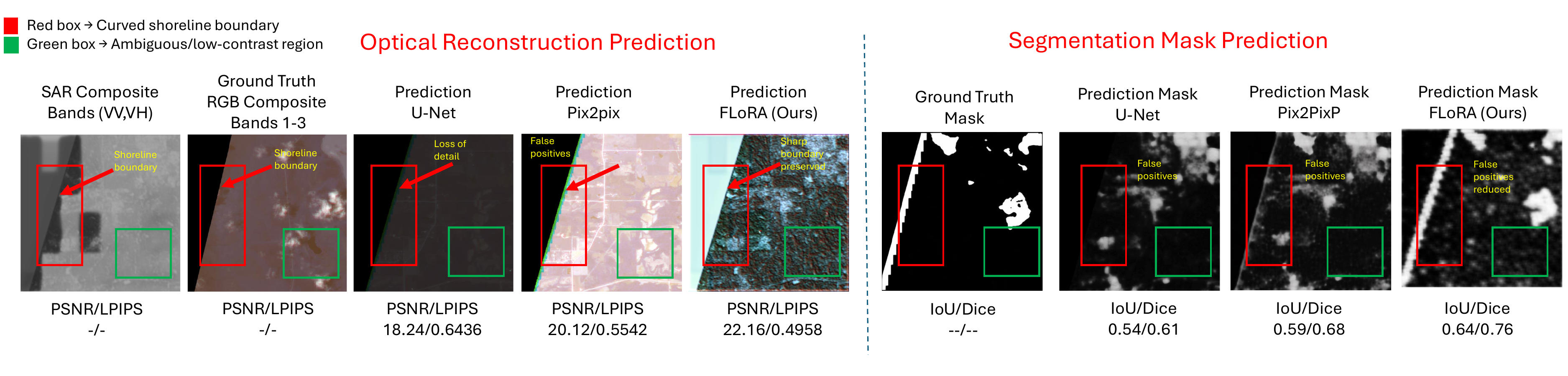}
  \caption{Visual comparison of SAR-to-Optical translation and flood-water segmentation on SEN1FLOODS11 \cite{bonafilia2020sen1floods11} dataset. Predicted masks are thresholded to binary (threshold = 0.5) for consistent comparison with ground-truth masks.}
  \label{fig:qualitative}
\end{figure*}

\begin{figure*}[t]
  \centering
  \includegraphics[width=\linewidth]{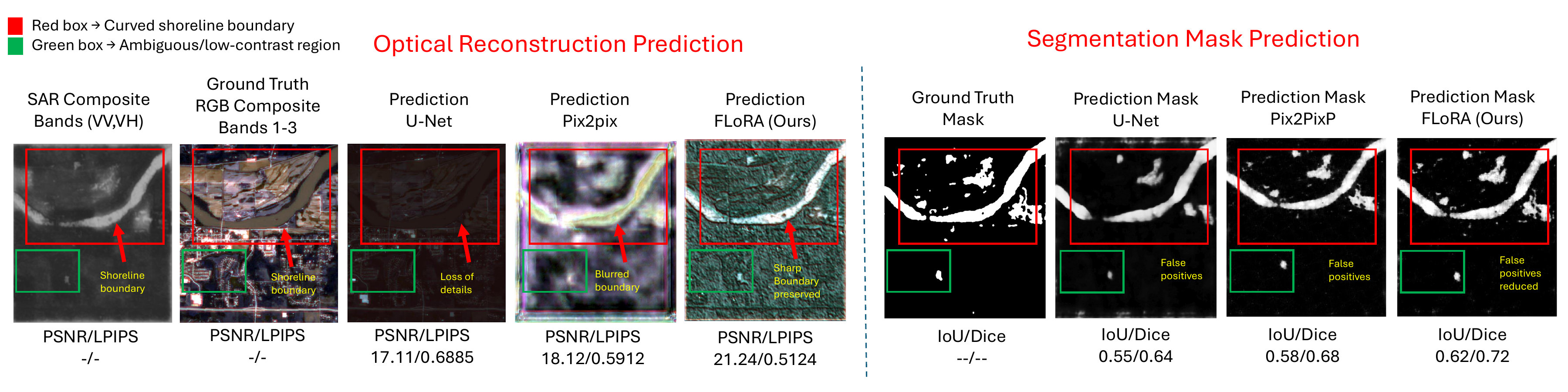}
  \caption{Visual comparison of SAR-to-Optical translation and flood-water segmentation on SEN12MS \cite{schmitt2019sen12ms} dataset. Predicted masks are thresholded to binary (threshold = 0.5) for consistent comparison with ground-truth masks.}
  \label{fig:qualitative}
\end{figure*}

\begin{figure*}[t]
  \centering
  \includegraphics[width=\linewidth]{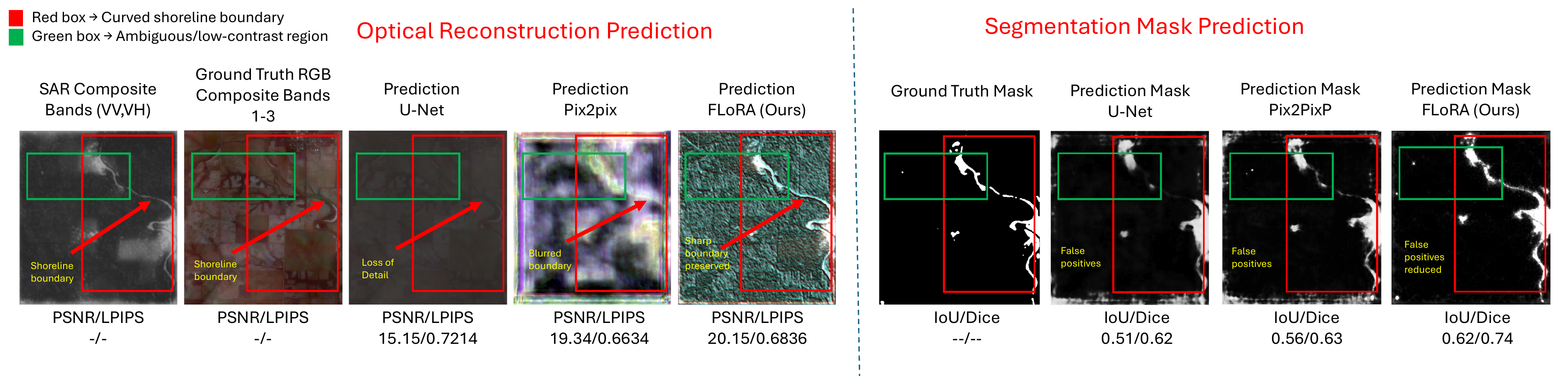}
  \caption{Visual comparison of SAR-to-Optical translation and flood-water segmentation on DEEPFLOOD \cite{fawakherji2025deepflood} dataset. Predicted masks are thresholded to binary (threshold = 0.5) for consistent comparison with ground-truth masks.}
  \label{fig:qualitative}
\end{figure*}

\raggedbottom

\subsection{Model Complexity vs. Reconstruction Performance}

Figure 6 presents the trade-off between reconstruction accuracy and model complexity for the evaluated methods on SEN1FLOODS11 \cite{bonafilia2020sen1floods11} dataset. While GAN- and transformer-based baselines improve performance over U-Net, they do so at the cost of increased parameter counts without proportional gains in PSNR.

In contrast, FLoRA achieves the highest reconstruction accuracy with a comparatively compact model size, occupying the most favorable position in the performance–complexity space. This demonstrates that the proposed fusion-aware latent reconstruction effectively balances accuracy and efficiency, making it well-suited for large-scale remote sensing applications.

\begin{figure}[t]
\centering
\includegraphics[width=0.95\linewidth]{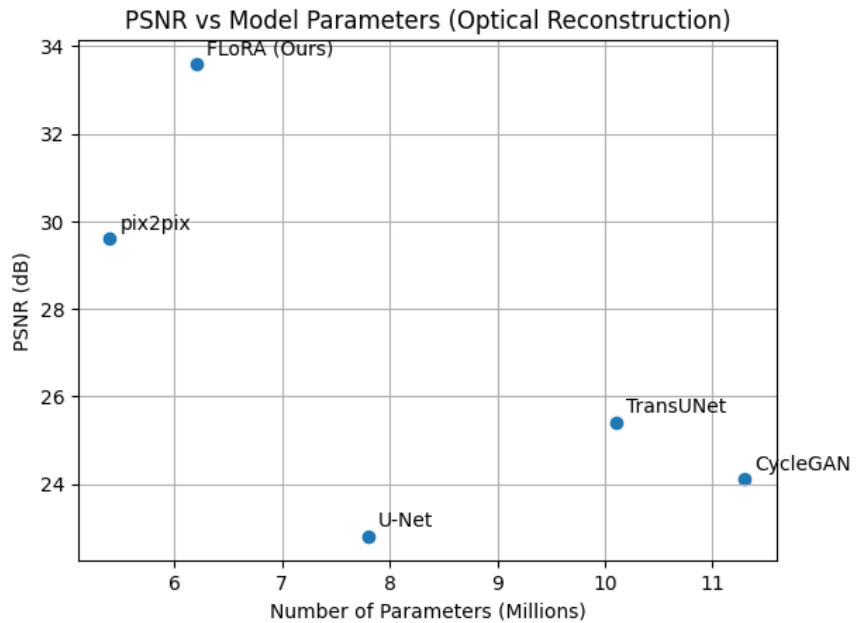}
\caption{Model Parameters vs Optical Resconstruction Performance on SEN1FLOODS11 \cite{bonafilia2020sen1floods11} dataset.}
\label{fig:confusion_matrix}
\end{figure}

\subsection{Performance and Efficiency Analysis}

Table VII summarizes reconstruction accuracy, segmentation performance, and computational complexity. While GAN- and transformer-based methods improve over U-Net, they require substantially higher FLOPs without proportional gains.

FLoRA achieves the highest PSNR and IoU while maintaining lower computational cost than both GAN- and transformer-based baselines, demonstrating an effective balance between accuracy and efficiency. This makes FLoRA well suited for large-scale and resource-constrained remote sensing applications.

\begin{table}[t]
\centering
\caption{Comparison of reconstruction accuracy, segmentation performance, and computational complexity. FLOPs and Multiply–Adds are reported for a single forward pass at a fixed input resolution.}
\begin{tabular}{lcccc}
\toprule
\textbf{Method} 
& \textbf{PSNR$\uparrow$} 
& \textbf{IoU$\uparrow$} 
& \textbf{FLOPs (G)$\downarrow$} 
& \textbf{Multi-Adds (G)$\downarrow$} \\
\midrule
U\mbox{-}Net \cite{ronneberger2015u}      
& 22.81 & 0.57 & 18.6 & 9.3 \\

CycleGAN \cite{zhu2017unpaired}           
& 24.12 & 0.63 & 42.8 & 21.4 \\

TransUNet \cite{zhu2017unpaired}          
& 25.38 & 0.61 & 55.6 & 27.8 \\

pix2pix \cite{isola2017image}             
& 29.56 & 0.66 & 34.1 & 17.0 \\

\textbf{FLoRA (ours)}                     
& \textbf{33.61} & \textbf{0.71} & \textbf{29.4} & \textbf{14.7} \\
\bottomrule
\end{tabular}
\label{tab:perf_flops}
\end{table}

\subsection{Ablation Study}
To analyze the contribution of each component, we performed controlled ablation experimets:

\subsubsection{Ablation study on different components of FLoRA}

The ablation results in Table VIII confirm that FiLM conditioning enhances spectral realism, teacher-guided fusion improves structural coherence, and gradient decoupling stabilizes multi-task optimization.

\begin{table}[t]
\centering
\caption{Ablation study on different components of FLoRA over the SEN1FLOODS11  \cite{bonafilia2020sen1floods11} Dataset.}
\begin{tabular}{lccc}
\toprule
\textbf{Configuration} & \textbf{PSNR$\uparrow$} & \textbf{SSIM$\uparrow$} & \textbf{IoU$\uparrow$} \\
\midrule
w/o FiLM Conditioning & 25.1 & 0.6311 & 0.66\\
w/o Optical Teacher Distillation & 24.7 & 0.6124 & 0.65\\
w/o Gradient Decoupling & 26.2 & 0.6728 & 0.68\\
\textbf{Full FLoRA} & \textbf{27.9} & \textbf{0.7212} & \textbf{0.71}\\
\bottomrule
\end{tabular}
\label{tab:ablation}
\end{table}

\subsubsection{Segmentation Accuracy and Confusion Analysis}
Figure 7 illustrates that FLoRA maintains low false alarms and misses for flooded vs. non-flooded pixels. Results on SEN12MS \cite{schmitt2019sen12ms} show slightly better class balance, indicating stronger generalization.

\begin{figure*}[t]
\centering
\includegraphics[width=0.8\linewidth]{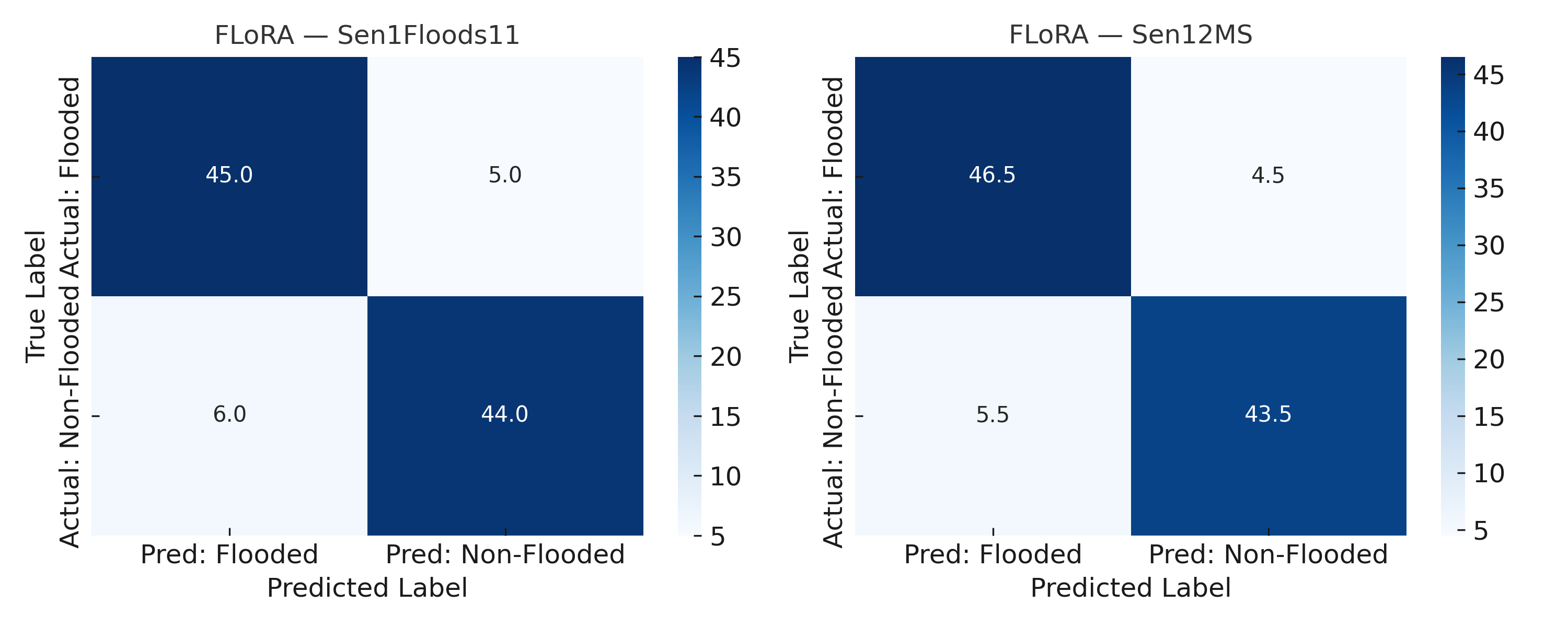}
\caption{Confusion matrices for FLoRA.}
\label{fig:confusion_matrix}
\end{figure*}

\subsubsection{Inference Time Analysis}

We report average inference time measured under identical hardware conditions (NVIDIA RTX 4090, batch size = 4, 256×256 input) for FLoRA in comparison to other SOTA methods. As shown in Table IX, FLoRA achieves favorable efficiency compared to transformer-based baselines while maintaining superior reconstruction and segmentation performance. This demonstrates that the proposed fusion-latent design improves accuracy without introducing prohibitive runtime overhead.
Table IX presents a joint comparison of reconstruction accuracy, segmentation performance, and computational efficiency on SEN1FLOODS11  \cite{bonafilia2020sen1floods11}. 
While U-Net \cite{ronneberger2015u} provides the fastest inference, its reconstruction and segmentation performance are limited. Transformer-based TransUNet \cite{zheng2025based} incurs significantly higher computational cost without proportional performance gains. 
FLoRA achieves the best trade-off, delivering superior reconstruction quality and segmentation accuracy with moderate runtime overhead.

\begin{table}[t]
\centering
\caption{Performance, model complexity, and inference efficiency comparison on the SEN1FLOODS11 \cite{bonafilia2020sen1floods11} dataset. 
Inference time is measured on an NVIDIA RTX 4090 GPU (batch size = 1, input size = 256$\times$256).}
\label{tab:performance_efficiency}
\resizebox{\columnwidth}{!}{
\begin{tabular}{lccccc}
\toprule
\textbf{Method} &
\textbf{PSNR$\uparrow$} &
\textbf{LPIPS$\downarrow$} &
\textbf{Dice$\uparrow$} &
\textbf{Params (M)$\downarrow$} &
\textbf{Inference Time (ms)$\downarrow$} \\
\midrule
U-Net \cite{ronneberger2015u}        & 22.81 & 0.7010 & 0.61 & 31.2  & 12.4 \\
CycleGAN \cite{zhu2017unpaired}      & 24.12 & 0.6740 & 0.66 & 47.5  & 17.2 \\
Pix2Pix \cite{isola2017image}        & 29.56 & 0.5870 & 0.74 & 54.7  & 18.6 \\
TransUNet \cite{zheng2025based}      & 25.38 & 0.6620 & 0.69 & 105.3 & 34.8 \\
\textbf{FLoRA (Ours)} 
              & \textbf{33.61} 
              & \textbf{0.4533} 
              & \textbf{0.79} 
              & \textbf{68.9} 
              & \textbf{22.7} \\
\bottomrule
\end{tabular}
}
\end{table}

\subsection{Analysis and Discussion}
\label{sec:analysis_discussion}

Beyond quantitative improvements, it is important to understand why FLoRA consistently outperforms baseline models across datasets.

\textbf{Perceptual Improvements (LPIPS):}  
FLoRA achieves consistent improvements in LPIPS, indicating improved perceptual realism rather than merely reduced pixel error. This improvement stems from the fusion-latent design. Windowed cross-attention aligns SAR structural cues with optical semantic representations at local scales, while FiLM conditioning reduces statistical mismatch between modalities. Additionally, FFT and edge-based losses enforce spectral consistency and boundary sharpness. Together, these mechanisms enable reconstruction of visually coherent textures that CNN-only and GAN-based baselines often fail to recover. Interestingly, FLoRA maintains superior LPIPS despite stronger GAN baselines, suggesting that fusion-latent guidance improves perceptual realism beyond adversarial learning alone.

\textbf{Component Contribution:}  
Ablation results (Table VIII) clarify the role of each component. Removing FiLM causes the largest degradation in reconstruction quality, confirming its importance in cross-modal statistical alignment. Eliminating teacher distillation reduces both reconstruction and segmentation performance, indicating that semantic guidance stabilizes the fusion representation. Gradient decoupling primarily improves multi-task stability; without it, competing objectives degrade segmentation accuracy. These findings suggest that cross-attention provides structural alignment, FiLM ensures feature calibration, and gradient decoupling balances joint optimization.

\textbf{Scenario-Based Advantages:}  
Qualitative results (Figures 3-5) show that FLoRA provides clearer benefits in challenging conditions. In urban scenes, strong SAR reflections often lead to false flood-water detections in baseline models; the gated fusion mechanism suppresses such errors by selectively integrating optical semantics. In vegetated or agricultural floodplains, where SAR backscatter is ambiguous, cross-modal alignment improves discrimination between inundated vegetation and wet soil. These cases highlight the advantage of structured fusion over simple feature concatenation.

The winter subset of SEN12MS was selected to provide a controlled experimental setting with stronger SAR-optical consistency and reduced seasonal domain shift and isolate cross-modal fusion behavior. While multi-season evaluation is important for assessing full generalization, such analysis is beyond the scope of this work and is left for future study. Seasonal variability introduces additional domain shift that is orthogonal to the primary focus of this study, namely cross-modal fusion and joint optimization.

Overall, the fusion-latent space enables semantically aligned and physically consistent representations that benefit both optical reconstruction and flood-water segmentation, explaining the consistent improvements observed across metrics and datasets.

\section{Conclusion}
\label{sec:conclusion}

We presented \textbf{FLoRA}, a unified cross-modal framework that jointly performs SAR-to-optical reconstruction and flood-water segmentation within a single model. By leveraging a fusion-latent representation guided by an optical teacher, FiLM-based feature alignment, and multi-scale cross-attention, FLoRA helps bridge the modality gap between SAR and optical imagery, improving both spectral realism and structural consistency. Extensive evaluations on SEN1FLOODS11, SEN12MS, and DEEPFLOOD demonstrate consistent gains over CNN-, GAN-, and transformer-based baselines in reconstruction quality, segmentation accuracy, and computational efficiency. These results highlight the effectiveness of fusion-aware latent learning for robust flood-water mapping, particularly in scenarios where optical data are limited or unavailable.

\section*{Acknowledgments}
This work is supported by NASA award 80NSSC23M0051 and NSF Award 2401942.

\begin{IEEEbiography}[{\includegraphics[width=1in,height=1.25in,clip,keepaspectratio]{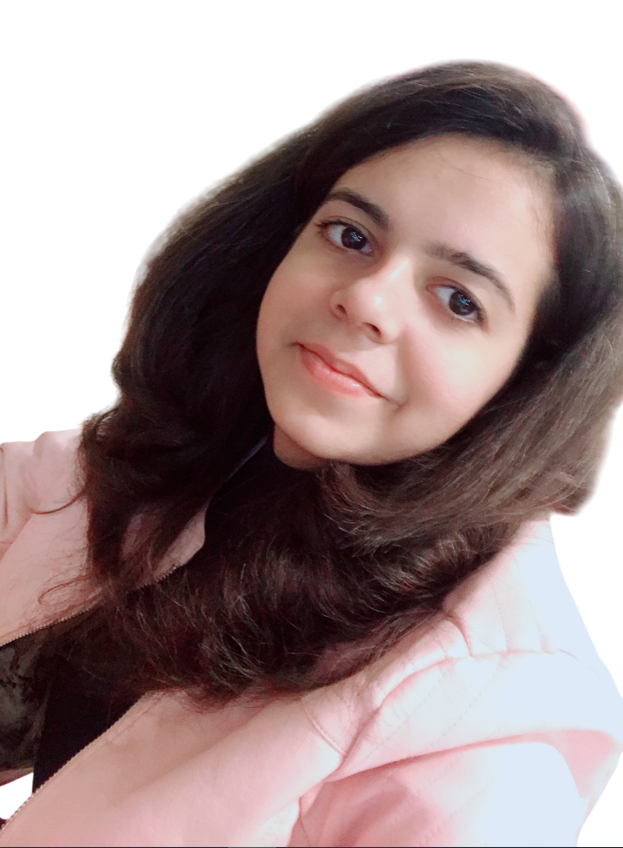}}]{JAGRATI TALREJA} (Graduate Member, IEEE) received the B.Tech. degree in Electronics and Communication Engineering from Pranveer Singh Institute of Technology, Kanpur, Uttar Pradesh, India, in 2019. She pursued and successfully completed a five-year integrated Ph.D. in Electrical Engineering at Chulalongkorn University, Bangkok, Thailand, in 2024.
\newline
Dr. Jagrati is currently a Postdoctoral Researcher at North Carolina A\&T State University, Greensboro, NC, USA. She is working on projects funded by NASA and the National Science Foundation (NSF), focusing on the development and application of deep learning algorithms and advanced geospatial techniques for analyzing satellite and UAV imagery in flood mapping and disaster assessment. Her research interests include Electrical Engineering, Remote Sensing, Satellite Image Processing, Neural Networks, and Machine Learning, with a particular emphasis on deep learning for image super-resolution, multimodal data fusion, and SAR-to-optical image translation and Quantum Computing. She has hands-on expertise in designing hybrid CNN–Transformer architectures, GAN-based frameworks, and transfer learning models for geospatial applications.
\end{IEEEbiography}

\begin{IEEEbiography}[{\includegraphics[width=1in,height=1.25in,clip,keepaspectratio]{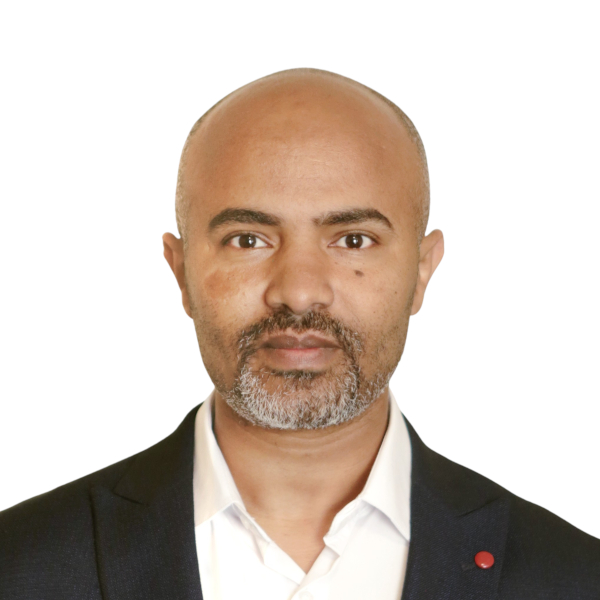}}]{TEWODROS SYUM GEBRE} holds a  B.S. in Hydraulic Engineering from Arba Minch University, Ethiopia, in 2009, an M.S. in Civil (Road and Transport) Engineering from Addis Ababa University, Ethiopia, in 2014, and a Ph.D  in Applied Science and Technology  at North Carolina A\&T State University, USA, in 2025. His expertise spans developing machine learning models for automating traffic management, image and both structure and unstructured data analysis, and system development. 
\newline
He is currently a Postdoctoral Researcher at North Carolina A\&T State University, in AI and ML applications for smart cities and Infrastructure Resilience. He worked on a Microsoft-funded project developing AI-based traffic monitoring systems using transformer models and UAV imagery. He is a former member of the NC-CAV Center, supported by the North Carolina Department of Transportation. He received the G. Herbert Stout Award for Best Student Paper at the North Carolina GIS (NCGIS) Conference and was awarded the Excellence Scholarship from 2021 to 2024. He also received Microsoft’s AFMR grant for the 2023–2024 period.
\end{IEEEbiography}

\begin{IEEEbiography}[{\includegraphics[width=1in,height=1.25in,clip,keepaspectratio]{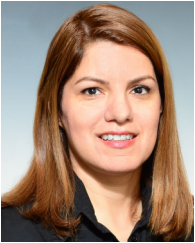}}]{LEILA HASHEMI-BENI} received the B.S.degree in civil-surveying engineering (geomatics) from the University of Isfahan, the M.S. degree in civil-surveying engineering (photogrammetry/remote sensing) from the University of Tehran, and the Ph.D. degree
in geospatial information system from Laval University. 
\newline
She is an Associate Professor and the Director of the NASA-funded Institute for Harnessing Data Science for Environment Management, North Carolina A\&T State University. She is a PI/a Co-PI on many projects supported by NASA, NSF, NOAA, Microsoft, North Carolina Collaboratory, and North
Carolina DoT. Her research on AI-based traffic monitoring systems using generative pre-trained transformer models and high-resolution UAV imagery is supported by Microsoft’ Accelerating Foundation Models Research Program. She is a Co-PI on the NC Transportation Center of Excellence on Connected and Autonomous Vehicle Technology (NC-CAV) funded by North Carolina Department of Transportation. Her research experience and interests include geospatial data science, UAV and satellite remote sensing, multi temporal and multisource data fusion and image classification, 3-D data modeling, automatic matching and change detection between various datasets, and developing GIS and remote sensing methodologies for environmental management. She has served as a proposal panelist and a reviewer for many U.S. and international funding organizations. She has served as the chair/the co-chair or as a scientific committee member for many national or international conferences/workshops. She is serving as the Co-Chair of LiDAR, Laser Altimetry and Sensor Integration Working Group, International Society of Photogrammetry and Remote Sensing (ISPRS).
\end{IEEEbiography}

\end{document}